\documentclass[runningheads]{llncs}
\usepackage{graphicx}
\usepackage{subcaption}
\usepackage{algorithm}
\usepackage{algorithmic}
\usepackage{cite}
\usepackage{amsmath}
\usepackage{amsfonts}
\usepackage{array}
\usepackage{url}
\usepackage[hidelinks]{hyperref}
\usepackage{cleveref}
\usepackage{adjustbox}
\usepackage{float}
\usepackage{textcomp}
\usepackage{comment}	
\usepackage{siunitx}    
\usepackage{amssymb}    
\usepackage{xcolor}		
\usepackage{caption}

\iftrue 
\graphicspath{{./Images/color/}}
\else 
\PassOptionsToPackage{monochrome}{xcolor}
\graphicspath{{./Images/grayscale/}}
\fi

\begin{document}
\newdimen\origiwspc
\newdimen\origiwstr
\origiwspc=\fontdimen2\font
\origiwstr=\fontdimen3\font

\title{Deep Learning Traversability Estimator for Mobile Robots in Unstructured Environments\thanks{This work has been carried out within the framework of the EUROfusion Consortium and has received funding from the Euratom research and training programme under grant agreement No 633053. The views and opinions expressed herein do not necessarily reflect those of the European Commission. The authors are grateful to the Autonomous Systems Group of RAL SPACE for providing the SEEKER dataset.}}
\titlerunning{DL Traversability Estimator for Mobile Robots}
%
\author{Marco Visca\inst{1} \and
Sampo Kuutti\inst{1} \and
Roger Powell\inst{2} \and
Yang Gao\inst{3} \and
Saber Fallah\inst{1}}
\authorrunning{M. Visca et al.}
%
\institute{Connected and Autonomous Vehicles Lab (CAV Lab), University of Surrey, Guildford, GU2 7XH, UK.
	\email{m.visca@surrey.ac.uk} \and
Cybernetics Group, Remote Applications in Challenging Environments, UK Atomic Energy Authority, Culham Science Centre, OX14 3DB\and
Space Technology for Autonomous and Robotic Laboratory (STAR LAB), Surrey Space Centre, University of Surrey, Guildford, GU2 7XH, UK
}
\maketitle              
\begin{abstract}
Terrain traversability analysis plays a major role in ensuring safe robotic navigation in unstructured environments. However, real-time constraints frequently limit the accuracy of online tests especially in scenarios where realistic robot-terrain interactions are complex to model. In this context, we propose a deep learning framework trained in an end-to-end fashion from elevation maps and trajectories to estimate the occurrence of failure events. The network is first trained and tested in simulation over synthetic maps generated by the OpenSimplex algorithm. The prediction performance of the Deep Learning framework is illustrated by being able to retain over $94\%$ recall of the original simulator at $30\%$ of the computational time. Finally, the network is transferred and tested on real elevation maps collected by the SEEKER consortium during the Martian rover test trial in the Atacama desert in Chile. We show that transferring and fine-tuning of an application-independent pre-trained model retains better performance than training uniquely on scarcely available real data.

\end{abstract}
\section{Introduction}
Autonomous traversability analysis of unstructured terrains is a crucial task in many sectors, such as rescue robots for disaster areas, agriculture, nuclear plants, and space exploration. The primary goal of traversability analysis is to ensure the safety of the robotic system and reduce its dependency on human control by autonomously assessing the surrounding terrain. Moreover, in contrast to navigation in structured environments, where a clear distinction between obstacle and non-obstacle is possible, unstructured natural terrains present continuous difficulty values which mostly depend on the specific robot mobility capabilities. This makes the definition of safe trajectories considerably more challenging as the algorithm has to take more numerous and complex metrics into account. On the other hand, real-time navigation requirements often impose stringent constraints on the overall software complexity.\\[-0.4cm]

In this context, several terrain analysis algorithms, which differently trade-off between accuracy and computational speed, have been proposed \cite{surrey721940}. Among them, square-grid cost maps based on geometric analysis are often considered the most successfully deployed on real systems \cite{GESTALT}\cite{exomars}. The main reason for their success is their implementation simplicity and relatively low computational workload. However, they often make use of overly conservative assumptions which could lead to sub-optimal navigation performance \cite{GESTALT}\cite{exomars}. Other works have proposed to use accurate physics-based simulators to assess the traversability of trajectories \cite{JPL_simulator}. However, in spite of their accuracy which allows to maximise the optimality of trajectory planning, their computational workload is often unbearable for on-board resources and real-time navigation.\\[-0.4cm]

In recent years, deep learning methods have gained an increasing popularity for their ability of extracting features from high-dimensional inputs and their efficient parallel computing \cite{lecun2015deeplearning}. In this context, deep learning has demonstrated remarkable capabilities to improve the autonomy of mobile robots \cite{JPL_classification}. Other works have proposed to exploit deep learning models to estimate mobile robot traversability metrics \cite{simone}\cite{8280544}. However, these methods often assess traversability over arbitrary-shaped patch of terrains (e.g. circular, or squared). Moreover, state-of-the-art deep learning methods often require substantial amounts of data to provide sensible predictions, while their availability is often limited for many robotic applications of interest.\\[-0.4cm]


In this paper, we propose a deep learning model to estimate traversability metrics from a simulator (Section \ref{sec:model_selection}). Our formulation has the advantage to explicitly address learning over feasible trajectories, thereby considering the robot mobility constraints and providing direct information in terms of trajectory planning.
Furthermore, we propose to address the problem of data scarcity by developing a synthetic dataset based on the OpenSimplex noise algorithm (Section \ref{sec:dataset_generation}). We show that, despite some degradation in performance, a model trained on the synthetic dataset can retain characteristics of generic unstructured terrains and, thus, be used as the baseline model of a real use-case scenario. We show evidence of this by transferring on real data from the SEEKER Martian rover test trial in the Atacama desert in Chile and comparing the synthetic model performance with training based solely on the limited amount of available real data (Section \ref{sec:results}).\\[-0.4cm]

\section{Traversability Prediction Model} \label{sec:model_selection}
We propose to formulate the traversability prediction as an image classification problem by using a standard Convolutional Neural Network (CNN) architecture. This choice is motivated by the well-demonstrated CNNs' capabilities to process and find patterns in high-dimensional spatial inputs \cite{cnn_advances}. Moreover, we propose to assess terrain traversability directly over feasible trajectories. In this way, failure prediction can be achieved by considering the actual robot mobility constraints, thereby increasing the accuracy of prediction. The remainder of this section illustrates the proposed methodology.

\subsection{Input Features}
\begin{figure}[t]
	\centering
	\includegraphics[width=0.44\linewidth]{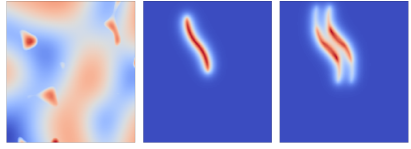}
	\caption{The three input layers. From the left: the terrain elevation map, the trajectory centre, and the wheel trace.}
	\label{fig:RGB_layers}
\end{figure}

To enable the use of CNN architectures, an approach is devised which gives the terrain and trajectory input features a three-channel image-like representation. Each channel is a \num{129}$\times$\num{129} grid, where each pixel position corresponds to an (x,y) coordinate with respect to the rover centre (between -4 and +4 meters) and the robot is assumed positioned in the centre of the map and oriented in the positive direction of the vertical axis. A visual description of the three channels is illustrated in Fig. \ref{fig:RGB_layers}. The three channels from left to right are: (1) the terrain elevations, where the value of each cell is the normalized z elevation value for that (x,y) coordinate, (2) the trajectory left from the robot on the map, where the channel has its peak (value of 1) at the robot centre and exponentially decreases to 0 at the wheel track, and (3) the trace left from the wheels on the map, where each trace has its peak at the wheel centre and exponentially decreases to 0 outside the wheel; furthermore, a higher value is given to the cells where both the front and rear wheels pass. 

In this way, the feature processing can be addressed on regions of the terrain of particular relevance to the failure prediction (i.e. the regions under the robot ride and the wheels) and directly over feasible trajectories. Hence, each \num{129}$\times$\num{129}$\times$\num{3} image represents one terrain-trajectory input feature that is fed to the neural network for traversability prediction.
\begin{figure}[t]
	\centering
	\includegraphics[width=0.75\linewidth]{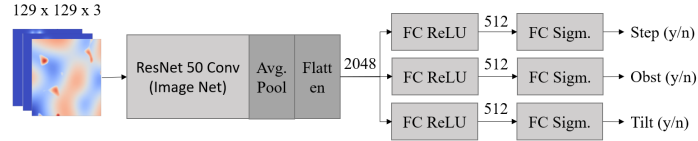}
	\caption{The proposed CNN architecture.}
	\label{fig:CNN_arch}
\end{figure}
\subsection{Network Architecture}\label{sec:network_architecture}
Figure \ref{fig:CNN_arch} illustrates the proposed neural network architecture. The ResNet50 network \cite{resnet} pre-trained on ImageNet \cite{imagenet} is chosen as the baseline of the prediction model. This choice is motivated by the remarkable performance demonstrated by Imagenet pre-trained residual networks as baseline architectures for transfer learning problem. Indeed, although our inputs are quite dissimilar from ImageNet images, exploiting pre-learned low level features (e.g. vertical or horizontal edges, which are common to all image classification problems) has proved to give faster convergence than training from scratch \cite{survey_transfer_learning}. Conversely, the original top Fully Connected (FC) layer is removed and replaced with three randomly initialized FC layers to learn the application-dependent features (one for each failure event as described in Section \ref{section:failure_events}). Each FC layer has 512 neurons and randomly initialized weights. Finally, three FC layers with sigmoid activation functions provide the failure predictions.

\subsection{Robot Model and Failure Events}\label{section:failure_events}
\begin{figure}[t]
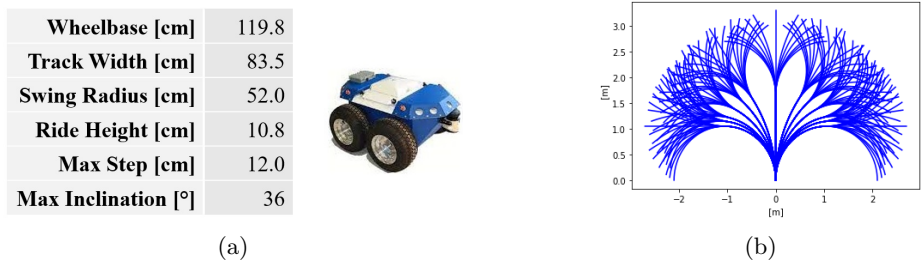

	\centering
	\begin{subfigure}[b]{0.5\linewidth}
		\includegraphics[width=\linewidth]{seekur_jr_v2}
		\caption{}
		\label{fig:robot}
	\end{subfigure}\hfill
	\begin{subfigure}[b]{0.35\linewidth}
		\includegraphics[width=\linewidth]{action_space_blue_0_v2.png}
		\caption{}
		\label{fig:action_space}
	\end{subfigure}
	\caption{(a) Seekur Jr. Robot (Courtesy of Generation Robots), and (b) portion of the robot action space for no initial point turn rotation.}
\end{figure}

A simplified kinematic robot model is developed in Python to emulate the robot navigation over unstructured terrains. The dimensions and the mobility capabilities of the robot model are selected according to the 4-wheel skid-steering Seekur Jr robot \cite{seekur}. Its main features are summarised in Fig. \ref{fig:robot}. Hence, each trajectory is defined according to the mobility capability of our robot as a combination of an initial point turn rotation (18 rotations for multiple of 20\textdegree including no rotation) followed by two arcs of length 1.65 meters and different radius (13 different possibilities). This leads to a total of 3042 possible trajectories 3.3 meters long each. Fig. \ref{fig:action_space} illustrates a portion of the action space for no initial rotation.

Three failure events are defined: \textit{step}, \textit{obstacle}, and \textit{tilt}. Failure for \textit{step} occurs when the differential elevation of the terrain underneath the robot wheels for two consecutive time steps is above the maximum traversable step of the robot. Failure for \textit{obstacle} occurs if one or more of the terrain elevation points underneath the robot base is higher than the robot ride height. Finally, failure for \textit{tilt} occurs if the inclination of the robot with respect to the vertical direction is above the maximum traversable inclination. To reduce the computational workload, the dynamics of the system are not taken into account. This is a reasonable assumption for robotics navigation at low speed (which could be the case in some realistic scenarios, such as planetary exploration and nuclear reactor maintenance) \cite{exomars}\cite{robot_nuclear}. Traverse is simulated by placing the robot on sequential trajectory points (equally spaced at \SI{6}{cm} intervals along each arc) and computing for each one of them the robot static pose and orientation and the elevation of the points under the rover base. Hence, the occurrence of the three failure events is recorded for each combination of terrain and trajectory.

\begin{algorithm}[t]
	\caption{Generating natural terrains with OpenSimplex}
	\label{alg:opensimplex}
	\begin{algorithmic}[1]
		\FOR {all x,y}
		\STATE $m = noise2d(x*\alpha_m,y*\alpha_m)*\beta_m + \gamma_m$ \label{eq:mountain}
		\STATE $p = (noise2d(x*\alpha_p,y*\alpha_p)*\beta_p + \gamma_p)^{\delta}$ \label{eq:plain}
		\STATE $w = intrp(noise2d(x*\alpha_w,y*\alpha_w)*\beta_w + \gamma_w, u, d)$ \label{eq:interp}
		\STATE $Z(x,y) = p*w + m*(1-w)$ \label{eq:terrain}
		\ENDFOR
		\RETURN $Z$
	\end{algorithmic}
	
\end{algorithm}

\begin{figure}[t]
	\centering
	\includegraphics[width=0.7\linewidth]{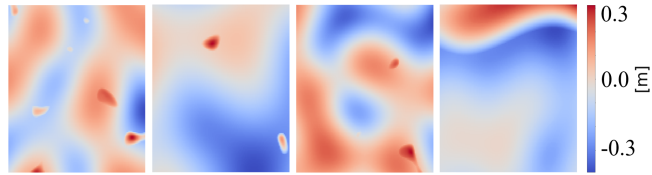}
	\caption{Examples of \SI{8}{m} $\times$ \SI{8}{m} maps generated using the OpenSimplex algorithm.}
	\label{fig:examples_terrains}
\end{figure}
\section{Dataset Generation} \label{sec:dataset_generation}
\subsection{OpenSimplex Synthetic Maps Generation}\label{sec:opensimplex}
To reduce the data scarcity problem of mobile robot applications, synthetic maps are generated using the OpenSimplex noise algorithm, a popular approach to generate realistic unstructured environments \cite{opensimplex_original}. In this work, the OpenSimplex Python API is used along with three filtering techniques to render realistic terrains. A description of the approach is illustrated in Algorithm \ref{alg:opensimplex}. The $noise2d$ function is the Python API which takes as input an (x,y) coordinate and outputs a number in [\num{-1},\num{1}] according to the OpenSimplex algorithm. Hence, additional heuristic parameters are used to filter the result of Opensimplex. Specifically, $\alpha_m$, $\alpha_p$, and $\alpha_w$ act on the noise frequency, $\beta_m$, $\beta_p$, and $\beta_w$ scale the output, while $\gamma_m$, $\gamma_p$, $\gamma_w$ offset the output. In this way, $\alpha_m$, $\beta_m$, and $\gamma_m$ are set in Line \ref{eq:mountain} to control the generation of obstacles. Line \ref{eq:plain} controls the generation of plain regions by using a smoothing coefficient $\delta \in$ [0,1] in addition to the $\alpha_p$, $\beta_p$, and $\gamma_p$ parameters. Line \ref{eq:interp} controls the interpolation between obstacles and plains, where $intrp$ is a function returning 1 if the first argument is larger than $u$, 0 if it is lower than $d$, or linearly interpolates between 0 and 1 otherwise. Finally, Line \ref{eq:terrain} combines the results of the previous three operations by interpolating between obstacles and plain regions and assigning the elevation value to the elevation matrix Z. The implementation of Algorithm \ref{alg:opensimplex}, with the parameters used in this paper, is made available at \textcolor{green}{\href{https://github.com/picchius94/unstruct_navigation.git}{UNSTR-NAV}}.  We remark that the process is fully automated and, by different tuning of the algorithm parameters, different terrain conditions can be achieved, such as rough, wavy, and smooth terrains, as well as mountains and depressions. Fig. \ref{fig:examples_terrains} illustrates some examples of generated maps.

\subsection{Dataset Collection and Training}
A total of 56840 synthetic elevation maps is generated with the method described in Section \ref{sec:opensimplex}. Then, the robot traverses each map with 3042 trajectories and collects failure events with the method described in Section \ref{section:failure_events}. The resulting dataset is composed of approximately 1.7e8 samples. The dataset is randomly divided among training (90\%), validation (8\%), and test (2\%) datasets. Moreover, since safe trajectories are considerably more numerous than failures for each terrain (90.4\% against 9.6\%), a reduced and better balanced subset is extracted for training and validation to avoid excessive bias in prediction (5.7e5 and 4.9e4 samples respectively). Conversely, all maps and trajectories of the test set are retained to assess final performance (3.4e6 samples).

The network is trained by means of supervised learning and binary cross-entropy loss function \cite{binary_cross}. The parameters used for training are: RMSprop optimizer, learning rate 1e-4, dropout 20\%, and L2 regularization 0.001. During the first epoch, only the 3 FC layers are trained, while the ResNet weights are kept frozen. Then, the whole network is unfrozen and trained for 10 epochs.

\section{Results} \label{sec:results}
\subsection{Prediction Performance - Synthetic Dataset}
The results of the trained model on the synthetic test dataset are illustrated in Table \ref{tab:synthetic_performance}. An overall accuracy of 98\% can be observed, with the accuracy of each failure event above 96\%. However, since safe and unsafe trajectories are extremely unbalanced in the test set (roughly 1e7 vs 4e5 samples respectively), accuracy by itself can not be considered as a representative metric. For this reason, recall, precision, and F1 score are used to provide a more informed representation of the actual model performance. Specifically, an overall high recall (94.4\%), and low precision (68.4\%) are observed, with a consequent F1 score of 79\%. This means that the network tends to be conservative, being able to correctly predict the majority of dangerous trajectories, but at a price of relatively high false-positive rate. A possible explanation for the low network precision could depend on the high sensitiveness of a correct prediction to small variations in the image. Indeed, even just one different pixel in the elevation map could lead the same trajectory to be safe or unsafe, making it a relatively challenging image classification problem.
Therefore, while the model could have successfully learned macro-associations of elevation points and trajectories to safe or unsafe areas, it might struggle to seize much more subtle local differences. However, the tendency to conservativeness is not excessively detrimental for the specific application of robotic navigation as long as safety is ensured. For instance, an increased rate of false alarms can be tolerated if it results in reliable identification of dangerous trajectories.
\begin{table}[t]
	\caption{Synthetic Dataset Model Performance}
	\label{tab:synthetic_performance}
	\begin{subtable}[t]{0.5\textwidth}
		\centering
		\subcaption{Confusion Matrices}
		\label{tab:test_results_cf}
		\begin{tabular}{ccc}
			Step & \textbf{Pred. Safe} & \textbf{Pred. Fail}\\
			\textbf{True Safe} & 3126951 & 113457 \\
			\textbf{True Fail} & 13615 & 201689 \\[0.05cm]
			Obstacle & \textbf{Pred. Safe} & \textbf{Pred. Fail}\\
			\textbf{True Safe} & 3211837 & 53842 \\
			\textbf{True Fail} & 9644 & 180389 \\[0.05cm]
			Tilt & \textbf{Pred. Safe} & \textbf{Pred. Fail}\\
			\textbf{True Safe} & 3425479 & 15806 \\
			\textbf{True Fail} & 163 & 14264 \\
			
		\end{tabular}
	\end{subtable}
	\begin{subtable}[t]{0.5\textwidth}
		\centering
		\subcaption{Classification Performance}
		\label{tab:test_results}
		\begin{tabular}{ccccc}
			
			& \textbf{Acc.} & \textbf{Recall} & \textbf{Prec.} & \textbf{F1 Score}  \\
			
			\textbf{Step} & $0.963$ & $0.937$ & $0.640$ & $0.760$\\
			
			\textbf{Obstacle} & $0.982$ & $0.950$ & $0.770$ & $0.850$\\
			
			\textbf{Tilt} & $0.995$ & $0.989$ & $0.474$ & $0.641$\\
			
			\textbf{Overall} & $0.980$ & $0.944$ & $0.684$ & $0.793$\\
			
		\end{tabular}
	\end{subtable}
	
\end{table}

\begin{figure}[t]
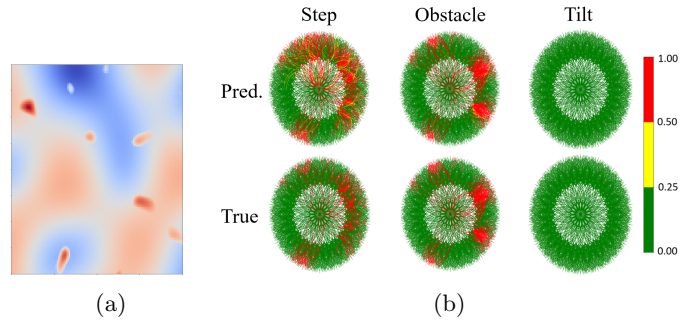

	\centering
	\begin{subfigure}[b]{0.22\linewidth}
		\includegraphics[width=\linewidth]{ex_map_v2}
		\caption{}
	\end{subfigure}
	\begin{subfigure}[b]{0.5\linewidth}
		\includegraphics[width=\linewidth]{34442_qualitative_results_v3}
		\caption{}
	\end{subfigure}
	\caption{Qualitative example of events prediction. (a) Synthetic elevation map, (b) prediction (top) and ground truth (bottom) of step, obstacle, and tilt probability.}
	\label{fig:qual_00}
\end{figure}

A qualitative example of the failure prediction is illustrated in Fig. \ref{fig:qual_00}. The image on the left represents the elevation map under analysis, while each subsequent image represents the robot action space (i.e. 3042 different trajectories from the map centre) with probability of failure occurrence encoded with 3 colours (\textbf{green}: less than 0.25, \textbf{yellow}: 0.25-0.5, \textbf{red}: more than 0.5). The network accurately predicts most of the dangerous trajectories due to obstacle, as wells as the absence of tilt failures. Conversely, some conservativeness can be observed for the step failure predictions. 
Nevertheless, trajectories laying in completely failure-free areas of the map are correctly predicted as safe which is a sign that the network has successfully learned to differentiate among macro distributions of elevation points.

The experiments are tested on a Intel Core i5 6500 Skylake and NVIDIA GeForce GTX 1050 graphic card. The resulting running time for a complete simulation of the 3042 trajectories is assessed at around 58s and 17s for the Python simulator and deep learning model respectively. Therefore, the deep learning model is able to reduce the computational time by approximately 70\%.

\begin{figure}[t]
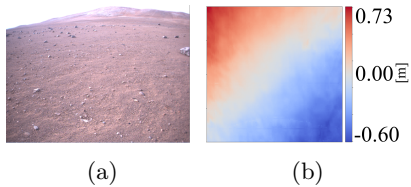

	\centering
	\begin{subfigure}[b]{0.21\linewidth}
		\includegraphics[width=\linewidth]{BEEX2_0_1338059364.3463_0_v2}
		\caption{}
		\label{fig:image_seeker}
	\end{subfigure}
	\begin{subfigure}[b]{0.22\linewidth}
		\includegraphics[width=\linewidth]{Figure_5_modified_v3}
		\caption{}
		\label{fig:map_seeker}
	\end{subfigure}
	
	\caption{(a) Image from the Atacama desert and (b) the extracted elevation map.}
	\label{fig:map_extraction}
\end{figure}

\subsection{Prediction Performance - Planetary Mission Use Case}
In this section, the traversability estimator is analysed on real unstructured terrains from the SEEKER Martian rover test trial in the Atacama desert in Chile \cite{seeker}. Fig. \ref{fig:image_seeker} shows an example of stereo camera image from which the test site elevation maps have been generated. From the SEEKER dataset, elevation maps from \num{10} traverses are selected for a total of \SI{1289}{m}. The real data are partitioned in \num{8} $\times$ \num{8} metres elevation maps to be consistent with the dimensionality of the input data accepted by our model. An example of extracted map from the SEEKER dataset can be observed in Fig. \ref{fig:map_seeker}. The final dataset is composed of \num{645} maps, which is approximately \SI{1.1}{\percent} of the synthetic dataset size. Moreover, data augmentation is performed to help reducing the data scarcity problem (by rotating each image by \num{90}, \num{180}, and \num{270} degrees). Hence, each sample is labelled according to the three failure events by running the traverse simulator.
Finally, one of the rover traverse is randomly selected and all its samples are removed from the training set to be used as the test set.

\begin{table}[t]
	\caption{Real Dataset Transferring Model Performance}
	\label{tab:real_performance}
	\begin{subtable}[t]{0.5\textwidth}
		\centering
		\subcaption{Confusion Matrices}
		\label{tab:test_results_2_cf}
		\begin{tabular}{ccc}
			Step & \textbf{Pred. Safe} & \textbf{Pred. Fail}\\
			\textbf{True Safe} & 132880 & 811 \\
			\textbf{True Fail} & 109 & 48 \\[0.05cm]
			Obstacle & \textbf{Pred. Safe} & \textbf{Pred. Fail}\\
			\textbf{True Safe} & 132622 & 521 \\
			\textbf{True Fail} & 269 & 436 \\[0.05cm]
			Tilt & \textbf{Pred. Safe} & \textbf{Pred. Fail}\\
			\textbf{True Safe} & 133735 & 113 \\
			\textbf{True Fail} & 0 & 0 \\
			
		\end{tabular}
	\end{subtable}
	\begin{subtable}[t]{0.5\textwidth}
		\centering
		\subcaption{Classification Performance}
		\label{tab:test_results_2}
		\begin{tabular}{ccccc}
			
			& \textbf{Acc.} & \textbf{Recall} & \textbf{Prec.} & \textbf{F1 Score}  \\
			\textbf{Step} & $0.993$ & $0.306$ & $0.056$ & $0.094$\\
			\textbf{Obst.} & $0.994$ & $0.618$ & $0.456$ & $0.525$\\
			\textbf{Tilt} & $0.999$ & - & - & -\\
			\textbf{Overall} & $0.995$ & $0.561$ & $0.251$ & $0.347$\\
			
		\end{tabular}
		
	\end{subtable}
\end{table}

First, the transferring performance on the SEEKER test set of our baseline model (i.e. pre-trained on the synthetic data but without further training on the real data) can be observed in Table \ref{tab:real_performance}. We observe that no failure for tilt is present in the real data. Therefore, we are limited in the evaluation of this event. However, the network is able to predict 99.9\% of the samples correctly as tilt safe. Furthermore, also the step and obstacle classes are extremely unbalanced towards safe trajectories. Similarly to the synthetic data, the accuracy is above 99\% (i.e. the model is able to classify nearly all the samples correctly as safe). Conversely, the performance for the failure prediction considerably drops both for the step and obstacle events. A possible explanation for this is that the synthetic data may not represent with sufficient realism many of the geometric distributions responsible for failure events in the real data. Specifically, the step event seems the most largely influenced both in terms of recall and precision (respectively 46\% and 6\%), while the obstacle event has been able to retain considerably better performance (recall of 62\% and precision of 46\%), which means that the network has learned to generalize more effectively to this type of failure in the real scenario.

\begin{figure}[t]
	\centering
	\includegraphics[width=0.9\linewidth]{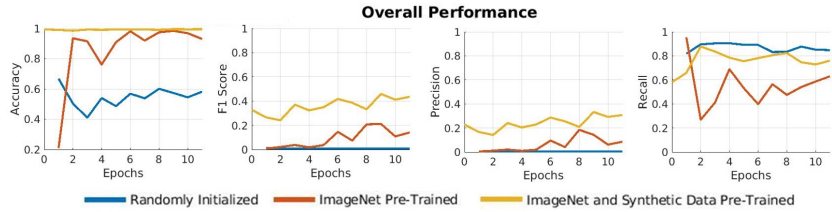}
	\caption{Overall test performance of differently pre-trained models while training on the real data. Epoch 0 on the yellow line is the transfer learning performance (i.e. our synthetic pre-trained baseline model before training on the real data).}
	\label{fig:training_comparisons}
\end{figure}
\begin{figure}[t]
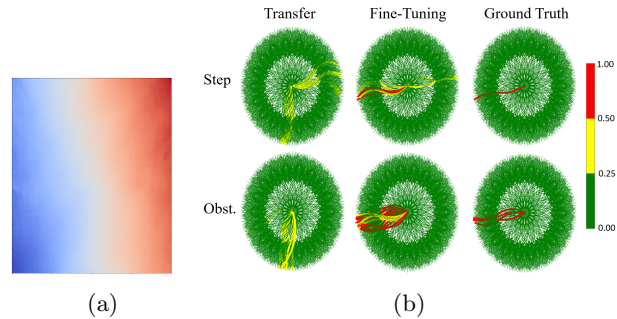

	\centering
	\begin{subfigure}[b]{0.2\linewidth}
		\includegraphics[width=\linewidth]{figure_real_example_v2}
		\caption{}
	\end{subfigure}
	\begin{subfigure}[b]{0.45\linewidth}
		\includegraphics[width=\linewidth]{qualitative_comparisons_tot_2_v3}
		\caption{}
	\end{subfigure}
	\caption{(a) Elevation map from SEEKER dataset, and (b) step and obstacle predictions before and after model fine-tuning compared with the ground truth.}
	\label{fig:qual_comparisons_real}
\end{figure}

Then, the performance of training on the SEEKER dataset is analysed for three differently pre-trained models: (1) a model with randomly initialized network parameters, (2) a model pre-trained on ImageNet only, and (3) our baseline model initialized with ImageNet weights and pre-trained on our synthetic dataset as described in Sections \ref{sec:model_selection} and \ref{sec:dataset_generation}. Hence, the three models are trained on the SEEKER training set for 11 epochs and the maximum F1 score on the test set is used as the convergence point of their performance. Fig. \ref{fig:training_comparisons} summarises our findings. The randomly initialized network fails to learn useful features in the dataset, resulting in poor performance. Meanwhile, the network pre-trained on ImageNet shows some initial improvement, learning useful features for its task, but overfits after 8 epochs at 21\% F1 score, resulting in 54\% recall and 14\% precision. Conversely, the model pre-trained on our synthetic dataset improves its performance more effectively when fine-tuning on the real dataset, resulting in a final F1 score, recall and precision of 46\%, 76\% and 31\%, respectively. This provides evidence of the improved capability of our baseline model to transfer features of relevance to the traversability analysis problem. Most importantly, this model is able to outperform the ImageNet model in terms of recall, demonstrating it has learned how to correctly classify failure events more reliably. In Fig. \ref{fig:qual_comparisons_real} a qualitative example is illustrated of the prediction capabilities of our baseline model before and after fine-tuning on the real data. As with the previous results, the transferred model is initially unable to correctly identify the dangerous trajectories both for the step and obstacle events while evidently improves after the model fine-tuning.

\section{Conclusion and Future work}\label{sec:conclusion}
This paper has investigated the use of deep learning as a traversability estimator for mobile robots in unstructured terrains. We provided insights on the benefits of the proposed method to predict the occurrence of failure events over feasible trajectories and at a fraction of the time of a sequential traverse simulator. We also showed that by generating a domain-independent synthetic dataset we can learn general features for traversability analysis. Then, by fine-tuning the learned model on the domain-specific real-world data, we can transfer the knowledge to enable the deep neural network to learn traversability analysis on datasets that would be excessively small to train on. Therefore, this technique enables more efficient learning for domains where large real world datasets are not available and where deep learning might not otherwise be a feasible solution due to the scarcity of data.

We are extending this work in several directions. While binary failure metrics could be sufficient to enforce safety in navigation, they cannot provide adequate information to perform optimal path planning. Future works may consider an extension of learning to include more complex continuous metrics. Finally, we discussed how the representativeness of the synthetic data could have a crucial impact during transfer learning. In future works, this could be addressed by generating more realistic synthetic data and by using algorithms specifically devised for efficient transfer learning (e.g. meta-learning).

%
%
%
 \bibliographystyle{splncs04}
 \bibliography{bibliography}

\begin{thebibliography}{10}
\providecommand{\url}[1]{\texttt{#1}}
\providecommand{\urlprefix}{URL }
\providecommand{\doi}[1]{https://doi.org/#1}

\bibitem{simone}
{Blacker}, P., {Bridges}, C.P., {Hadfield}, S.: Rapid prototyping of deep
  learning models on radiation hardened cpus. In: 2019 NASA/ESA Conference on
  Adaptive Hardware and Systems (AHS). pp. 25--32 (July 2019)

\bibitem{8280544}
Chavez-Garcia, R.O., Guzzi, J., Gambardella, L.M., Giusti, A.: Learning ground
  traversability from simulations. IEEE Robotics and Automation Letters
  \textbf{3}(3),  1695--1702 (2018). \doi{10.1109/LRA.2018.2801794}

\bibitem{GESTALT}
{Goldberg}, S.B., {Maimone}, M.W., {Matthies}, L.: Stereo vision and rover
  navigation software for planetary exploration. In: Proceedings, IEEE
  Aerospace Conference. vol.~5, pp.~5--5 (2002)

\bibitem{cnn_advances}
Gu, J., Wang, Z., Kuen, J., Ma, L., Shahroudy, A., Shuai, B., Liu, T., Wang,
  X., Wang, G., Cai, J., et~al.: Recent advances in convolutional neural
  networks. Pattern Recognition  \textbf{77},  354--377 (2018)

\bibitem{resnet}
He, K., Zhang, X., Ren, S., Sun, J.: Deep residual learning for image
  recognition. CoRR  \textbf{abs/1512.03385} (2015)

\bibitem{JPL_simulator}
Helmick, D., Angelova, A., Matthies, L.: Terrain adaptive navigation for
  planetary rovers. Journal of Field Robotics  \textbf{26} (04 2009)

\bibitem{robot_nuclear}
{Iqbal}, J., {Tahir}, A.M., {ul Islam}, R., {Riaz-un-Nabi}: Robotics for
  nuclear power plants — challenges and future perspectives. In: 2012 2nd
  International Conference on Applied Robotics for the Power Industry (CARPI).
  pp. 151--156 (Sep 2012)

\bibitem{lecun2015deeplearning}
LeCun, Y., Bengio, Y., Hinton, G.: Deep learning. Nature  \textbf{521}(7553),
  436--444 (2015)

\bibitem{JPL_classification}
{Ono}, M., {Fuchs}, T.J., {Steffy}, A., {Maimone}, M., {Yen}, J.: Risk-aware
  planetary rover operation: Autonomous terrain classification and path
  planning. In: 2015 IEEE Aerospace Conference. pp. 1--10 (March 2015)

\bibitem{seekur}
ROBOTS, G.: {Seekur Jr mobile robot},
  \url{https://www.generationrobots.com/en/402399-robot-mobile-seekur-jr.html}

\bibitem{binary_cross}
Ruby, U., Yendapalli, V.: Binary cross entropy with deep learning technique for
  image classification. International Journal of Advanced Trends in Computer
  Science and Engineering  \textbf{9} (10 2020)

\bibitem{imagenet}
Russakovsky, O., Deng, J., Su, H., Krause, J., Satheesh, S., Ma, S., Huang, Z.,
  Karpathy, A., Khosla, A., Bernstein, M.S., Berg, A.C., Li, F.: Imagenet large
  scale visual recognition challenge. CoRR  \textbf{abs/1409.0575} (2014)

\bibitem{surrey721940}
Sancho-Pradel, D., Gao, Y.: A survey on terrain assessment techniques for
  autonomous operation of planetary robots. JBIS - Journal of the British
  Interplanetary Society  \textbf{63}(5-6),  206 -- 217 (May 2010)

\bibitem{opensimplex_original}
Spencer, K.: {Introducing OpenSimplex Noise},
  \url{https://www.reddit.com/r/proceduralgeneration/comments/2gu3e7/like_perlins_simplex_noise_but_dont_like_the/ckmqz2y/}

\bibitem{survey_transfer_learning}
Tan, C., Sun, F., Kong, T., Zhang, W., Yang, C., Liu, C.: A survey on deep
  transfer learning. In: Artificial Neural Networks and Machine Learning --
  ICANN 2018. pp. 270--279. Springer International Publishing, Cham (2018)

\bibitem{exomars}
Winter, M., Rubio, S., Lancaster, R., Barclay, C., Silva, N., Nye, B., Bora,
  L.: Detailed description of the high-level autonomy functionalities developed
  for the exomars rover. In: 14th Symposium on Advanced Space Technologies in
  Robotics and Automation (06 2017)

\bibitem{seeker}
Woods, M., Shaw, A., Tidey, E., Van~Pham, B., Simon, L., Mukherji, R.,
  Maddison, B., Cross, G., Kisdi, A., Tubby, W., Visentin, G., Chong, G.:
  Seeker—autonomous long-range rover navigation for remote exploration.
  Journal of Field Robotics  \textbf{31}(6),  940--968 (2014)

\end{thebibliography}

%
%
%
%
\end{document}